\documentclass[10pt,twocolumn,letterpaper]{article}

\usepackage{cvpr}              
\usepackage{verbatim}
\usepackage{amsfonts,amssymb} 
\usepackage{multirow}
\usepackage{graphicx}
\usepackage{caption}
\usepackage{lipsum}
\usepackage[accsupp]{axessibility}
\usepackage[pagebackref,breaklinks,colorlinks,citecolor=cvprblue]{hyperref}

\usepackage[dvipsnames]{xcolor}

\definecolor{cvprblue}{rgb}{0.21,0.49,0.74}

\def\paperID{7072} 
\def\confName{CVPR}
\def\confYear{2024}

\title{TexVocab: Texture Vocabulary-conditioned Human Avatars}

\author{Yuxiao Liu$^1$ \quad  Zhe Li$^2$ \quad  Yebin Liu$^2$ \quad  Haoqian Wang$^1$\\
$^1$Shenzhen International Graduate School, Tsinghua University    \quad $^2$Tsinghua University \\
}

\begin{document}


\twocolumn[{
\renewcommand\twocolumn[1][]{#1}
\maketitle
\begin{center}
    \captionsetup{type=figure}
    \includegraphics[width=0.96\linewidth]{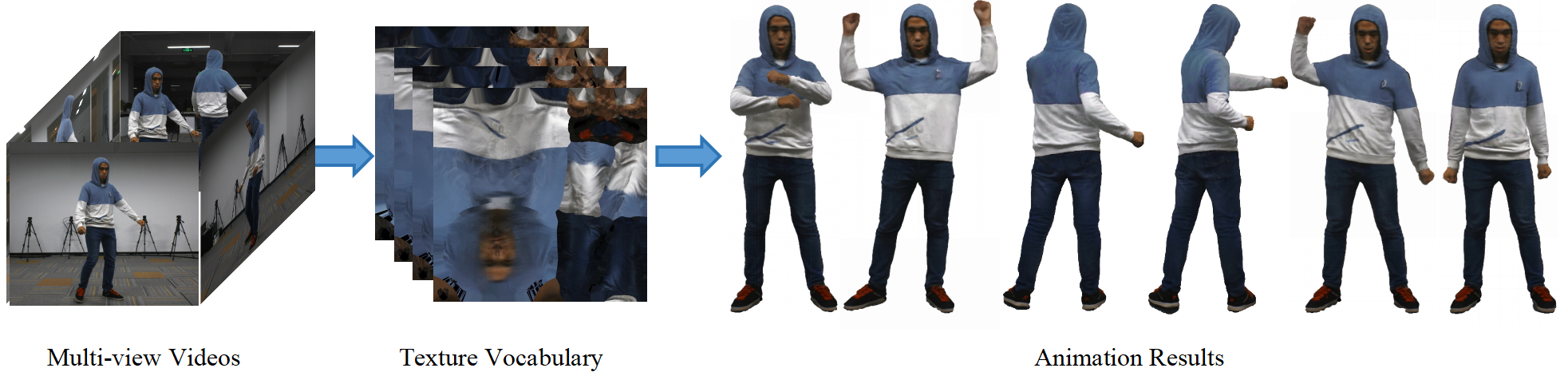}
    \captionof{figure}{Overview of TexVocab. Given multi-view RGB videos of one character, we construct a texture vocabulary, and create realistic animatable human avatars. }
    \label{fig:fig1}
\end{center}
}]

\begin{abstract}

To adequately utilize the available image evidence in multi-view video-based avatar modeling, we propose TexVocab, a novel avatar representation that constructs a texture vocabulary and associates body poses with texture maps for animation. 
Given multi-view RGB videos, our method initially back-projects all the available images in the training videos to the posed SMPL surface, producing texture maps in the SMPL UV domain. 
Then we construct pairs of human poses and texture maps to establish a texture vocabulary for encoding dynamic human appearances under various poses.
Unlike the commonly used joint-wise manner, we further design a body-part-wise encoding strategy to learn the structural effects of the kinematic chain.
Given a driving pose, we query the pose feature hierarchically by decomposing the pose vector into several body parts and interpolating the texture features for synthesizing fine-grained human dynamics.
Overall, our method is able to create animatable avatars with detailed and dynamic appearances from RGB videos, and the experiments show that our method outperforms state-of-the-art approaches. The project page can be found at \href{https://texvocab.github.io/}{\tt https://texvocab.github.io/}.

\end{abstract}    
\section{Introduction}
\label{sec:intro}

\hspace{0.422cm}
Animatable human avatar modeling attracts a lot of attention due to its enormous potential in AR/VR applications including games, movies and holoportation. 
Animatable human avatars usually take the skeletal body pose as the input signal and output the pose-conditioned dynamic human appearances.
However, how to effectively learn the mapping between the driving signals and dynamic appearances is still full of challenges.

Previous works~\cite{li2022tava,zheng2022structured} usually directly map the pose input, e.g., the pose vectors, to the human appearances using a conditional neural radiance field (NeRF)~\cite{mildenhall2021nerf} represented by an MLP.
However, the pose input does not involve any information about dynamic human appearances, so it remains difficult for NeRF MLPs to regress high-fidelity dynamic details solely from the pose input.
Although some works~\cite{peng2021animatable,wang2022arah,li2023posevocab} propose to auto-decode~\cite{park2019deepsdf} latent embeddings to encode the dynamic appearances at the input end of NeRF, they still suffer from the limited representation ability of global codes \cite{peng2021animatable} or feature lines \cite{li2023posevocab}, resulting in blurry synthesized avatars.

On the other hand, image-based reconstruction methods like pixelNeRF~\cite{yu2021pixelnerf} and SparseFusion~\cite{zhou2023sparsefusion} have already proved that taking pixel-aligned features as the input of NeRF can significantly improve the quality and details for static scene rendering.
Inspired by these image-conditioned representations, we propose TexVocab, a texture vocabulary that adequately utilizes explicit image evidence to guide the implicit conditional NeRF to learn the dynamics from expressive texture conditions.
To associate the multi-view images with the dynamic human body, we back-project all the available images of corresponding training poses to the posed SMPL surface and transform them to the SMPL UV domain like NeuralActor~\cite{liu2021neural}, obtaining a set of texture maps.
Compared with other 3D representations like feature lines~\cite{li2023posevocab} and feature planes~\cite{chan2022efficient,chen2022tensorf}, the 2D texture maps can compactly cover the whole 2D manifold of the human body while avoiding excessive memory cost.
Besides, the texture maps provide pixel-aligned features that already involve detailed human appearances so that they can serve as effective conditions to decode fine-grained details.

Our objective is to not only reconstruct all the frames in the training dataset, but also synthesize high-quality dynamic human appearances under novel poses.
Therefore, it is necessary to bridge the human poses and texture maps for animation.
Drawing inspiration from PoseVocab~\cite{li2023posevocab}, we sample key human poses and associate them with corresponding texture maps.
Unlike PoseVocab which constructs a separate vocabulary for each joint of the SMPL model~\cite{2015SMPL}, we observe that the dynamic wrinkles of clothed humans are not individually influenced by the relative rotations of a single joint.
Instead, they are typically influenced by the kinematic chain of a body part, such as the entire right arm.
To this end, we propose to decompose all the SMPL skeletons into several body parts, then sample key body parts and assign corresponding texture maps to them.
Such a body-part-wise decomposition explicitly models the structural motion of each body part, enabling better generalization to novel poses.


In summary, our technical contributions are below:
\begin{itemize}[left=0.422cm]
    \item TexVocab, a new avatar representation that constructs a texture vocabulary to leverage expressive texture conditions for high-quality avatar modeling.
    \item A body-part-wise encoding method that not only disentangles the effects of different joints on the dynamic appearance, but also retains the structural effects of the kinematic chain, enabling better pose generalization. 
    \item Experiments demonstrate that our method can create higher-fidelity avatars with dynamic pose-dependent details compared to other SOTA approaches.
\end{itemize}

\section{Related Works}
\hspace{0.422cm}{\bf Implicit Neural Representations.}
In the last few years, implicit scene representation is becoming increasingly popular as it can produce impressive results both in geometry~\cite{chen2019learning,mescheder2019occupancy,park2019deepsdf} and appearance~\cite{lombardi2019neural, mildenhall2021nerf,sitzmann2019scene,thies2019deferred} modeling. 
Recent works show that implicit representations like occupancy networks~\cite{mescheder2019occupancy}, signed distance fields (SDF)~\cite{park2019deepsdf} and neural radiance fields (NeRF)~\cite{mildenhall2021nerf} lead to higher-resolution and topology-free 3D scene modeling compared with explicit ones, e.g., voxels~\cite{maturana2015voxnet,qi2016volumetric,song2016deep}, points~\cite{qi2017pointnet,qi2017pointnet++} and meshes~\cite{bronstein2017geometric,guo20153d,wang20183d}.
In particular, NeRF~\cite{mildenhall2021nerf} shows impressive rendering quality and good differentiable properties, attracting much attention in static scene rendering~\cite{garbin2021fastnerf,liu2020neural,yu2021plenoctrees,muller2022instant,hu2023tri,chen2022tensorf}. 
Many other works add the time dimension to extend NeRF to dynamic scene modeling~\cite{gao2021dynamic,gafni2021dynamic,park2021nerfies,tretschk2021non,shao2023tensor4d,icsik2023humanrf}. 
However, the motivation of these works is to reconstruct the dynamic scene, object or character in each frame of the given video without considering generating dynamic appearances under novel conditions, e.g., body poses.
While our goal is to not only reconstruct the 3D human under the training poses, but also to synthesize dynamic details under unseen poses.

{\bf Geometric Avatar Modeling.} Geometric based methods aim at training a pose-conditioned human model based on observations from geometric data like scans~\cite{saito2021scanimate,ma2021power,chen2021snarf,li2022avatarcap} or depth images~\cite{wang2021metaavatar,burov2021dynamic,dong2022pina}. 
Scan-based methods like SCANimate~\cite{saito2021scanimate}, SCALE~\cite{ma2021scale}, POP~\cite{ma2021power} and SNARF~\cite{chen2021snarf} adopt the SMPL \cite{2015SMPL} poses or position maps as the pose conditions to learn the pose-dependent geometric surface.
FITE~\cite{lin2022learning} and GeoTexAvatar~\cite{li2022avatarcap} render positional maps of posed SMPL models for the spatial continuity of the learned pose-dependent warping field. 
MetaAvatar~\cite{wang2021metaavatar}, PINA~\cite{dong2022pina} and DSFN~\cite{burov2021dynamic} learn the pose-dependent dynamics from depth sequences.
However, these methods require 3D data for training the avatar, limiting their applications to more accessible RGB videos.

\begin{figure*}
  \centering
  \includegraphics[width=0.93\linewidth]{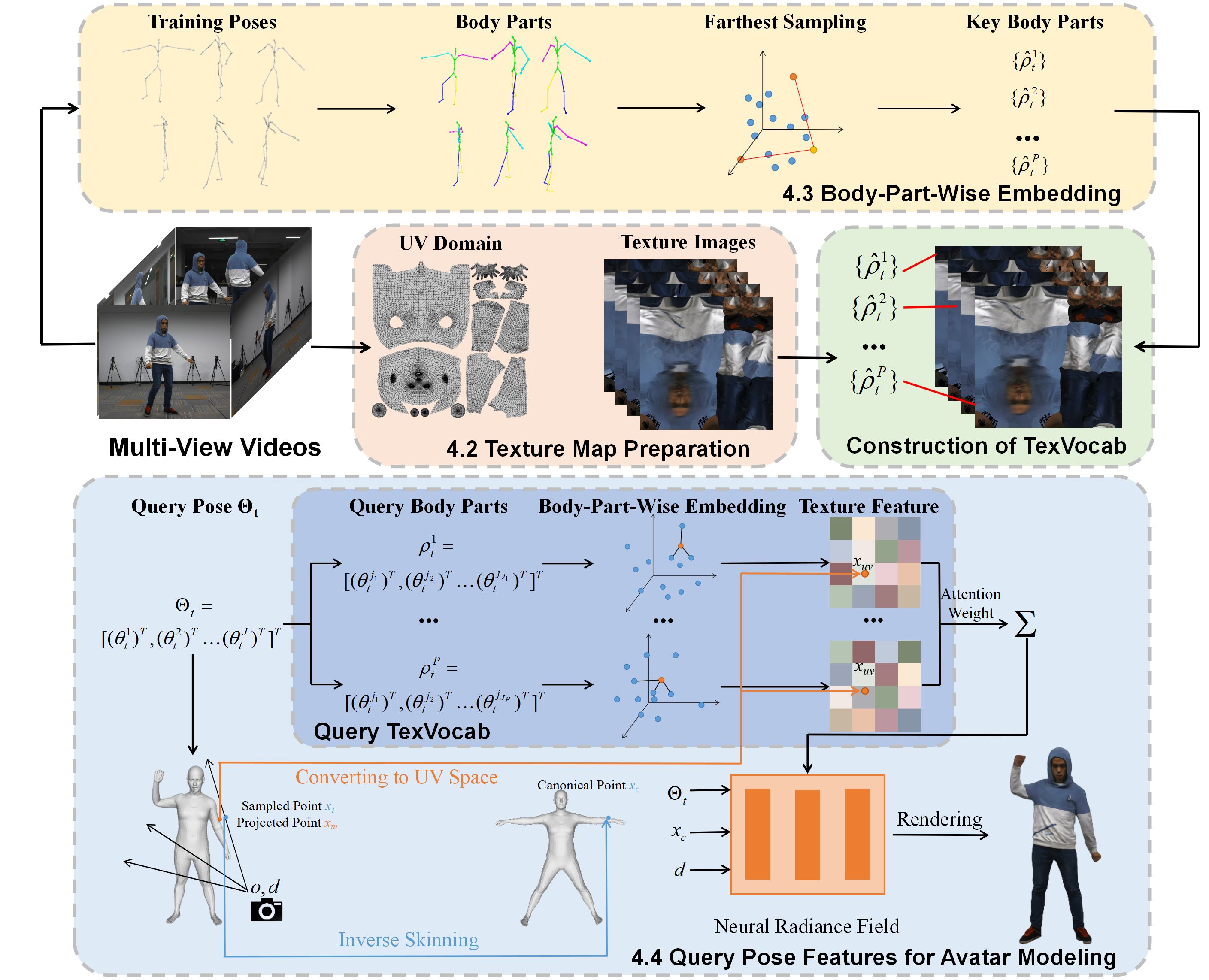}
   \caption{{\bf Framework of TexVocab}. We first construct TexVocab by decomposing SMPL poses into body parts, sampling key body parts and gathering corresponding texture maps. Then given a query pose and a 3D coordinate, we decompose the pose into body parts, interpolate key body parts and sample texture maps as the pose conditioned feature. We finally utilize NeRF represented as an MLP to decode the dynamic character and render human appearance with detailed pose-dependent dynamics.
   }
   \label{fig:methodfig}
\end{figure*}

{\bf RGB Video-based Avatar Modeling.}
On the other end of the spectrum, lots of works focus on creating animatable textured avatars from RGB videos~\cite{li2022tava,jiang2022selfrecon,zheng2022structured,zheng2023avatarrex,peng2021animatable,wang2022arah,li2023posevocab,liu2021neural,jiang2023instantavatar}.
Many approaches directly map the SMPL-derived inputs like pose vectors to the human appearances using a conditional NeRF~\cite{mildenhall2021nerf} to decode the dynamic character.
Specifically, TAVA~\cite{li2022tava} learns the non-rigid warping fields and shading effects directly conditioned on SMPL pose vectors. 
SelfRecon~\cite{jiang2022selfrecon} learns the canonical template mesh and pose-dependent deformation from a self-rotating video. 
SLRF~\cite{zheng2022structured} and AvatarRex~\cite{zheng2023avatarrex} sample nodes attached to SMPL, define a set of local radiance fields and learn the mapping from SMPL pose vectors to node translations and dynamic appearances.
However, the pose input does not contain any information about dynamic human appearances, so it remains difficult for MLPs to predict dynamics among various poses, thus limiting the quality of avatars.
AniNeRF~\cite{peng2021animatable}, ARAH~\cite{wang2022arah}, NeuralBody~\cite{peng2021neural} TotalSelfScan~\cite{dong2022totalselfscan} and PoseVocab~\cite{li2023posevocab} auto-decode~\cite{park2019deepsdf} latent embeddings to encode the dynamic appearances with per-frame latent code or joint-structured feature lines. 
Although adding extra embeddings variables like global codes~\cite{peng2021animatable} and feature lines~\cite{li2023posevocab} offloads the network, the low capability and poor representation ability of these embeddings still limits the avatar quality.
NeuralActor~\cite{liu2021neural} predicts pose-dependent texture maps from SMPL normal maps through the vid2vid model~\cite{wang2018video} using ground-truth texture maps as the monitoring signals, extracts and feeds textural feature to an MLP for decoding the dynamic human appearances.
TexDVA~\cite{remelli2022drivable} and LookingGood~\cite{martin2018lookingood} take 2 or 3 RGB images as the driving signals for synthesizing dynamic human appearances. 
Stylepeople~\cite{grigorev2021stylepeople} uses simple neural rendering instead of NeRF which allows to render photo-realistic 2D images of individuals in baggy clothes in different poses.

{\bf Avatar Modeling with Simulation.}
Unlike the data-driven methods mentioned above, another line of methods use physical simulation for dynamic garment reconstruction to model the clothed human avatars. Pioneer learning-based garment simulation methods~\cite{pan2022predicting,patel2020tailornet,santesteban2019learning,vidaurre2020fully} use pre-defined simulators and pre-generated data, while the deep learning framework does not contain any physical models. 
PBNS~\cite{bertiche2020pbns} and SNUG~\cite{santesteban2022snug} incorporate physical constraints into dynamic cloth simulation and generate realistic simulation results using unsupervised training.
Caphy~\cite{su2023caphy} proposes to optimize the parameters of the garment physics priors to obtain better physical properties from the 3D scans instead of using the given fixed physical parameters of the fabric. 
Compared with these works, our method is data-driven without the requirement of complicated simulation.
\section{Preliminary}
\hspace{0.422cm}Given a set of multi-view RGB videos of a performer with $T$ frames captured by $N$ synchronized cameras, we aim at training a model that can output high-fidelity dynamic appearances animated by skeletal poses. We denote the RGB sequences as $\{I_t^n| 1\leq n \leq N, 1\leq t \leq T\}$. We assume the access to the body poses of all the frames, denoted as $\Theta_t \in \mathbb{R}^{J\times3},1\leq t \leq T$, where $J$ denotes the number of joints of the human body. We assume that the former $T_1$ poses are used for training, while the rest poses are for testing. 
Similar to other avatar representations~\cite{peng2021animatable,te2022neural}, we utilize linear blend skinning (LBS) to transform sampled points from the observation space of pose $\Theta_t$ to the canonical space. The sampled point and the canonical point can be denoted as $x_t$ and $x_c$ respectively. Then we represent the canonical 3D character as a pose-conditioned neural radiance field (NeRF)~\cite{mildenhall2021nerf} that takes a canonical 3D coordinate $x_c$, a viewing direction $d$, and the positional conditioned pose feature $f(\Theta_t, x_c)$ as the input:
\begin{equation}
	g(\gamma_x(x_c), \gamma_d(d), f(\Theta_t, x_c)) = (\sigma_t(x_c), c_t(x_c))
 \label{eqA}
\end{equation}
where $\gamma_x$ and $\gamma_d$ is the positional encoding~\cite{tancik2020fourier} for spatial location and viewing direction. $\sigma_t$ and $c_t$ denote the output density and color conditioned on the pose $\Theta_t$. Following NeRF~\cite{mildenhall2021nerf}, we render the images using output density values and color values.

\section{Method}
\subsection{Overview}
\hspace{0.422cm}$f(\Theta_t, x_c)$ in Eq.~\eqref{eqA} plays an important role as it provides pose conditions for the MLP to decode the dynamic 3D character. Pioneer works like AniNeRF~\cite{peng2021animatable}, ARAH~\cite{wang2022arah} and PoseVocab~\cite{li2023posevocab} assign global latent codes or joint-structured feature lines as the pose conditions. However, these representations produce limited animation results because of the low representation ability of global codes and feature lines. To provide the pose-conditioned features with higher quality, we propose TexVocab, a novel method that adequately exploits explicit image evidence to guide the neural networks to learn the pose-dependent dynamic details. The framework of our approach is shown in Fig.~\ref{fig:methodfig}.
First, we prepare texture maps, decompose all the training poses and sample key body parts to construct TexVocab. Then given a SMPL pose and a 3D position, we query key body parts using K nearest neighbors (KNN), interpolate, and sample texture maps as the pose-conditioned feature according to the UV coordinate of the 3D position. Finally, we decode the dynamic 3D character using NeRF which is represented as an MLP.

\subsection{Texture Map Preparation}
\label{sec:32}
\hspace{0.422cm}
To utilize all the image evidence more efficiently, we propose to gather all the available training views to a particular UV domain and acquire texture maps. 

\begin{figure}
  \centering
 \includegraphics[width=0.96\linewidth]{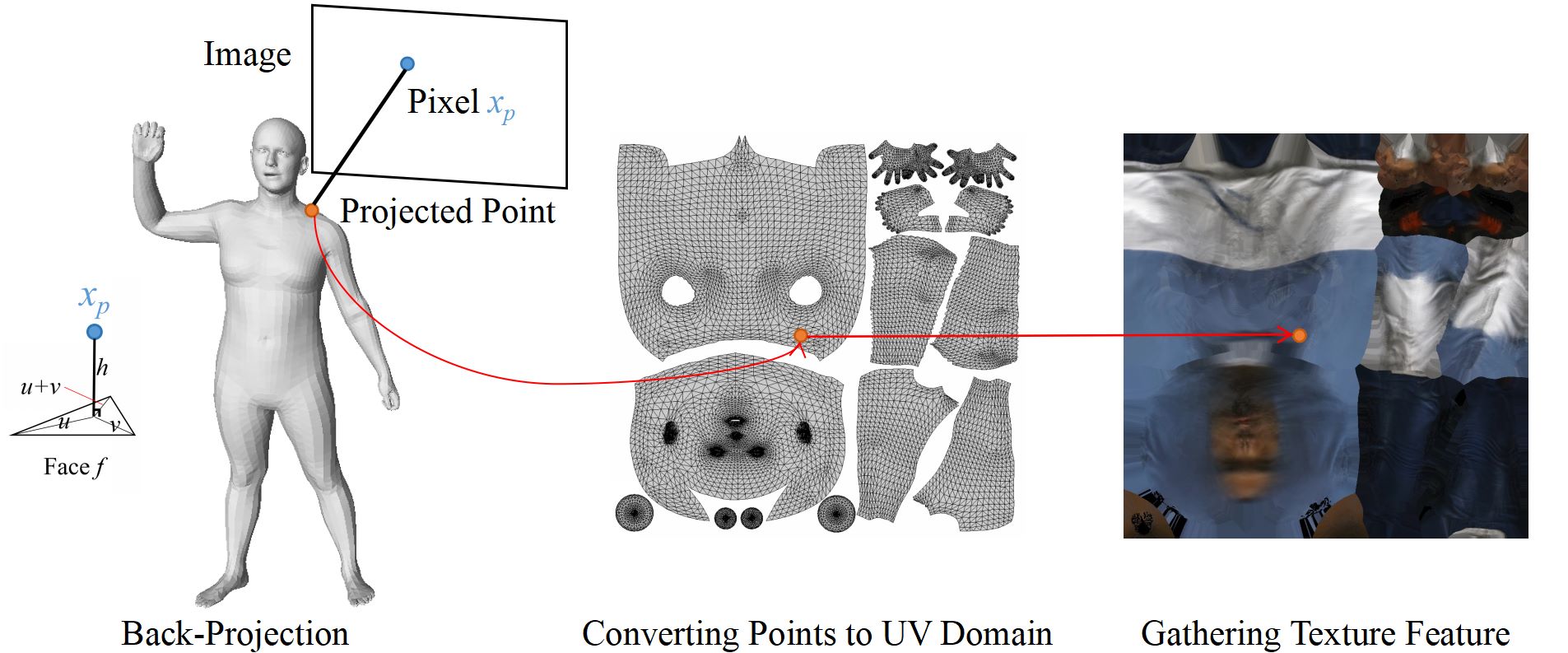}

   \caption{Overview of texture map preparation. First, we back-project all the available pixels to the posed SMPL mesh. Then we convert the projected points on the SMPL mesh to a particular UV domain. Finally, we gather and average all the available pixels, and obtain texture maps based on multi-view images.}
   \label{fig:projection}
\end{figure}

For each training pose $\Theta_t$, $1 \leq t \leq T_1$, as shown in Fig.~\ref{fig:projection},
we gather the available views and back-project every pixel $x_p$ to a posed SMPL mesh $(\mathcal{V}_t, \mathcal{F})$,:
\begin{equation}
    find (u^*, v^*, f^*) \,\, s.t. \,\, \Vert x_p - \mathcal{P}(\mathcal{B}_{u,v}(\mathcal{V}_{t[\mathcal{F}(f)]}))\Vert_2^2= 0
 \label{eqNA0}
\end{equation}
where $\mathcal{V}_t \in \mathbb{R}^{N_V\times3}$ and $\mathcal{V}_t \in \mathbb{R}^{N_F\times3}$ denotes the vertices and faces, $N_V$ and $N_F$ are the number of vertices and faces of the SMPL mesh respectively. $1 \leq f \leq N_F$ is the triangle index, $\mathcal{V}_{t[\mathcal{F}(f)]}$ is the three vertices of the triangle $\mathcal{F}(f)$, $(u,v): u, v, u+v \in [0, 1]$ represent the barycentric coordinates on the face, $\mathcal{B}_{u,v}(\cdot)$ is the barycentric interpolation function, and $\mathcal{P}(\cdot)$ stands for perspective projection.

Then we transfer the projected point $(u^*, v^*, f^*)$ to a particular UV domain based on the SMPL-defined UV parameterization $\mathcal{A} \in \mathbb{R}^{N_F \times 3 \times 2}$ which maps 3D mesh surface points to a 2D UV plane. Once all the image pixels have been mapped, we average it across all the available views to gather the texture image $U_t$. Finally, we extract the feature map $F_t$ of the texture image $U_t$ using a convolutional neural network (CNN) similar to PixelNeRF~\cite{yu2021pixelnerf}.

\subsection{Body-Part-Wise Embedding}
\label{sec:33}
\begin{figure}
  \centering
  \includegraphics[width=0.96\linewidth]{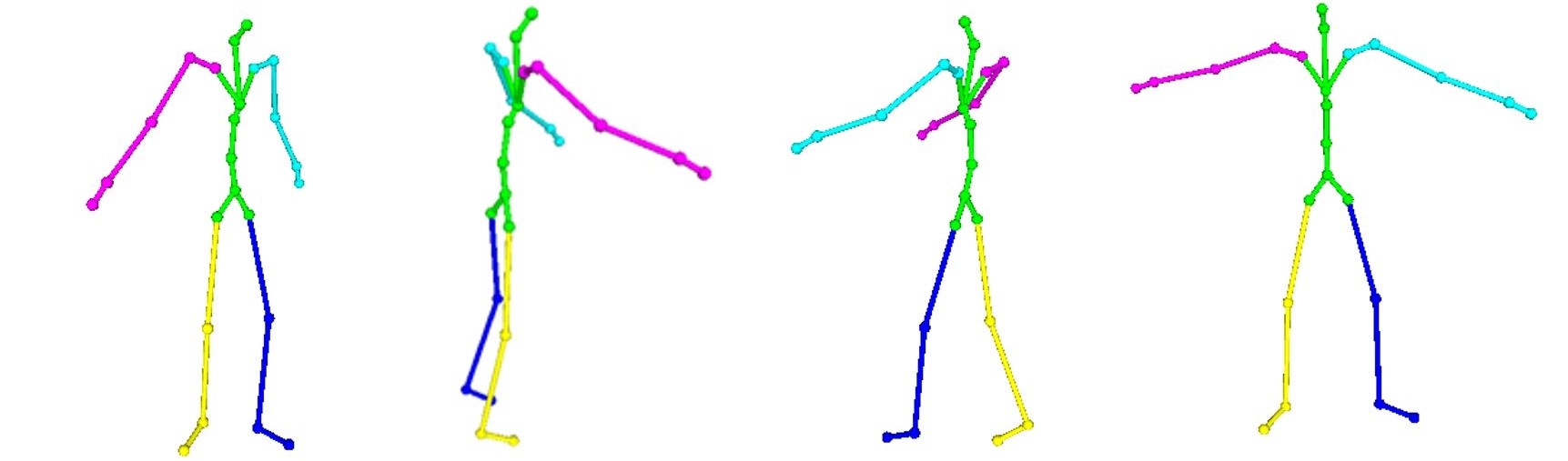}

   \caption{We decompose the SMPL skeletons into several body parts. Joints with the same color belong to the same body part.}
   \label{fig:bodypart}
\end{figure}

\hspace{0.422cm} We observe that the dynamic wrinkles of clothed humans are not influenced by the rotation of a single joint individually, but governed by a kinematic chain of a body part. To this end, we propose to divide poses into several body parts, which can not only disentangle the effects of different joints on the dynamic appearances but also retain topology information among the kinematic chains. Specifically, as shown in Table~\ref{tab:bodypart} and Fig.~\ref{fig:bodypart}, we divide the SMPL pose with $J=24$ joints into $P=5$ body parts, and each body part contains a particular set of joints. We denote the $p$-th body part of the $t$-th frame as

\begin{equation}
    \rho_t^p=[(\theta_t^{j_1})^T,(\theta_t^{j_2})^T,\dots,(\theta_t^{j_{J_p}})^T]^T
\end{equation}
here $\theta_t^j \in so(3)$ denotes the $j$-th joint rotation of pose $\Theta_t$, $1 \leq p \leq P$ is the rank of the body part, and $J_p$ denote the number of joints contained in the $p$-th body part.

After the division, we sample key body parts and assign embeddings to them. Given the $p$-th body parts $\{ \rho_t^p|1\leq t \leq T_1 \}$ of the training poses, we first sample $M$ key body parts via farthest point sampling. The distance metric between two body parts $\rho_1^p, \rho_2^p$ is calculated as the sum of the distance~\cite{huynh2009metrics} of each joint rotation in the $p$-th body part:
\begin{equation}
    d(\rho_1^p, \rho_2^p)=\sum_{k=1}^{J_p}( 1-|q(\theta_1^{j_k})^Tq(\theta_2^{j_k})| )
    \label{eq:distance}
\end{equation}
where $q(\cdot)$ is a function that maps an axis-angle vector to a unit quaternion. We denote the sampled $M$ key body parts as $\{ \hat{\rho}_m^p|1 \leq m \leq M\}$. For each sampled key body part $\hat{\rho}_t^p$, we assign the corresponding texture image feature $\hat{F_t}$ to it. Notice that we assign the same texture map to different key body parts which belong to the same pose.

So far, we have constructed M pairs of keys and values for each body part based on the training poses. For each key body part, we can find the corresponding textural feature. These body-part-wise pose embeddings construct TexVocab and serve as discrete samples in the continuous pose feature space in the following query.

\begin{table}
  \centering
  \begin{tabular}{@{}cc@{}}
    \toprule
    body part rank & Joints contained in the body part \\
    \midrule
    1(main body) & 0,1,2,3,6,9,12,15 \\
    2(left leg) & 4,7,10 \\
    3(right leg) & 5,8,11 \\
    4(left arm) & 13,16,18,20,22\\
    5(right arm) & 14,17,19,21,23\\
    \bottomrule
  \end{tabular}
  \caption{The division of the body parts. To maintain the information of the kinematic chains, we decompose SMPL skeletons into body parts instead of joints.}
  \label{tab:bodypart}
\end{table}

\subsection{Query Pose Features for Avatar Modeling}
\label{sec:34}
\hspace{0.422cm}Given a pose vector $\Theta_t=[(\theta_t^1)^T,(\theta_t^2)^T,\dots,(\theta_t^J)^T]^T$ and a 3D position $x_t$ sampled in the observation space of pose $\Theta_t$, we query the pose feature based on the constructed TexVocab in Sec.~\ref{sec:32} and Sec.~\ref{sec:33}.

{\bf Query Body Parts.} Given a pose $\Theta_t$, we first decompose it to several body parts denoted as $\{ \rho_t^p|1\leq p\leq P\}$. For the $p$-th query body part $\rho_t^p$, we search for K nearest body parts $\{ \hat{\rho}_k^p|1 \leq k \leq K \}$ according to Eq.~\eqref{eq:distance}. Then we interpolate the corresponding texture maps $\{ \hat{F}_k|1 \leq k \leq K \}$ according to the distance between body parts and make normalization:
\begin{equation}
    F_{p,t}=\sum\nolimits_{k=1}^K \hat{F}_k \cdot \dfrac{ w(\rho_t^p,\hat{\rho}_k^p)}{\sum\nolimits_{k=1}^K w(\rho_t^p,\hat{\rho}_k^p)} 
    \label{eq:F_interpolate}
\end{equation}
\begin{equation}
    w(\rho_t^p,\hat{\rho}_k^p)=J_p-d(\rho_t^p,\hat{\rho}_k^p)
\end{equation}
where $w(\rho_t^p,\hat{\rho}_k^p)$ is the weight that varies inversely to the distance between body parts.

{\bf Texture Map Sampling.} 
Given a 3D position $x_t$ in the observation space of $\Theta_t$, we project it to SMPL surface:
\begin{equation}
	(u^*, v^*, f^*)= {\arg\min_{u,v,f} \Vert x_p - \mathcal{B}_{u,v}(\mathcal{V}_{t[\mathcal{F}(f)]})\Vert_2^2}
 \label{eqNA1}
\end{equation}
and then we convert the projected point $x_m=(u,v,f)$ to the 2D coordinate $x_{uv}$ in UV domain according to the fixed UV parameterization matrix $\mathcal{A}$ defined in Sec.~\ref{sec:32}. 
For each weighted feature map $F_{p,t}$, we sample the feature $h(x_{uv}, F_{p,t})$ using bilinear interpolation. Moreover, following SCANimate~\cite{saito2021scanimate}, we also apply a skinning-weight-wise attention scheme on the feature of the $p$-th body part to reduce spurious correlations, which can be denoted as
\begin{equation}
    \Omega(x_c, p)=\dfrac{\sum\nolimits_{i=1}^{J_p}\omega(x_c, j_i)}{\sum\nolimits_{j=1}^J\omega(x_c, j)}
\end{equation}
where $\omega(x_c, j)$ is the pre-defined influence weight of the $j$-th joint on the corresponding canonical point $x_c$. Finally, we gather the sampled feature as
\begin{equation}
    f(\Theta_t, x_c)=\sum\nolimits_{p=1}^{P}\Omega(x_c, p)\cdot h(x_{uv}, F_{p,t})
\end{equation}

With the sampled pose feature $f(\Theta_t, x_c)$, we can feed it along with the pose vector $\Theta_t$, the viewing direction $d$ and the 3D coordinate $x_c$ into the network described as Eq.~\eqref{eqA} to decode NeRF that represents the dynamic 3D character.

{\bf Discussion.}
For the pixel $x_p$ and the query 3D coordinate $x_t$, we both project them to the posed SMPL mesh and convert them to a particular UV domain. These guarantee the alignment of the sampled textural feature. The texture maps are gathered from available views, which involve detailed human appearances and can provide effective conditions to decode fine-grained details.

Also, notice that the inverse skinning often includes residuals when sampled points are not on the SMPL mesh. To further ensure the alignment of the sampled textural feature, we do not use the canonical coordinate $x_c$, but use $x_t$ in the observation space of pose $\Theta_t$ instead. 

\begin{table*}[htb]
  \centering
\begin{tabular}{c|cccc|cccc}
\hline
\multirow{2}{*}{Method} & \multicolumn{4}{c|}{Training Poses} & \multicolumn{4}{c}{Novel Poses} \\ \cline{2-9} 
                 & PSNR$\uparrow$ & SSIM$\uparrow$ & LPIPS*$\downarrow$ & FID$\downarrow$ & PSNR$\uparrow$ & SSIM$\uparrow$ & LPIPS*$\downarrow$ & FID$\downarrow$ \\ \hline
                TAVA~\cite{li2022tava}  &  25.57   & 0.9624 & 29.58 & 64.59 &   26.61  & 0.9597 & 31.14 & 79.96 \\
                ARAH~\cite{wang2022arah}   & 22.78 & 0.9335 & 77.93 & 126.69 &   22.13  & 0.9241 & 93.06 &  113.88  \\
                AniNeRF~\cite{peng2021animatable}  & 25.19 & 0.9592 & 31.21 &   81.57 & 23.85 & 0.9486 & 32.07 & 95.41 \\
                PoseVocab~\cite{li2023posevocab}  & 34.06 & 0.9852 & 14.43 &    22.88 & 30.13 & 0.9806 & 16.32 & 28.10 \\
                Ours  & {\bf36.52}  & {\bf0.9896} & {\bf10.83} & {\bf12.31} & {\bf32.09}  & {\bf0.9832} & {\bf13.40} & {\bf18.79}   \\ \hline
\end{tabular}

  \caption{Quantitative comparisons against TAVA, ARAH, AniNeRF, and PoseVocab on sequence ``subject00" of THUman4.0 dataset. We evaluate the numerical results of each method on both training poses and novel poses. Here LPIPS* = 1000 $\times$ LPIPS.}
  \label{tab:quan-t4}
\end{table*}

\begin{table}[htb]
\centering
\begin{tabular}{cccc}
\hline
Method & PSNR$\uparrow$ & LPIPS$\downarrow$ & FID$\downarrow$ \\ \hline
NeuralActor~\cite{liu2021neural} & 23.531 & 0.066 & 19.714 \\
Ours & {\bf26.325} & {\bf0.017} & {\bf17.836}  \\ 
 \hline
\end{tabular}

\caption{Quantitative comparisons against NeuralActor on sequence ``S2" of ``DeepCap" dataset. Results of NeuralActor are borrowed from~\cite{liu2021neural}.}
\label{tab:quan_351}
\end{table}

\subsection{Training}
\hspace{0.422cm}We use pre-trained resnet34~\cite{he2016deep} as the backbone of the CNN that extracts the features of texture image, and the parameters are fixed during the training stage. 
Also, we do not regress the density value $\sigma_t$ in Eq.~\eqref{eqA} directly. Instead, we output SDF value $s_t$ and convert it to density value following VolSDF~\cite{yariv2021volume}. The total loss $\mathcal{L}$ includes the color loss, the mask loss, the eikonal loss, and the perceptual loss, which is defined as:
\begin{equation}
	\begin{split}		\mathcal{L}&=\lambda_{color}\mathcal{L}_{color}+\lambda_{mask}\mathcal{L}_{mask}\\&+\lambda_{perceptual}\mathcal{L}_{perceptual}+\lambda_{eikonal}\mathcal{L}_{eikonal}
	\end{split}
 \label{eq:loss}
\end{equation}
where $\lambda_{color}$, $\lambda_{perceptual}$, $\lambda_{mask}$ and $\lambda_{eikonal}$ stand for the loss weights. $\mathcal{L}_{color}$ measures the MSE between the rendered and ground-truth pixel colors, $\mathcal{L}_{mask}$ is an MAE loss which supervises the occupancy values of the rendered pixels, and $\mathcal{L}_{eikonal}$ is the Eikonal loss encouraging the geometry fields to approximate a true signed distance function~\cite{yariv2021volume}.
The perceptual loss $\mathcal{L}_{perceptual}$ is widely used in NeRF training, which leads to better recovery of high-frequency details like the clothed wrinkles and thin lines~\cite{zhang2018unreasonable}. We choose VGGNet as the backbone to calculate the learned perceptual image patch similarity (LPIPS).

\section{Experiments}

\subsection{Datasets and Metrics.}
\hspace{0.422cm}{\bf Datasets.} We use 6 sequences of multi-view videos for experiments. 3 sequences with 24 views are from THUman4.0 dataset~\cite{zheng2022structured}, 2 sequences with 21 or 23 views are from ZJU-MoCap dataset~\cite{peng2021animatable} and 1 with 11 views is from DeepCap dataset~\cite{habermann2020deepcap}. All the datasets provide parameters of cameras. Deepcap and THuman4.0 provide SMPL-X~\cite{pavlakos2019expressive} registrations, and ZJU-MoCap provides SMPL~\cite{2015SMPL}.
We split each sequence into 2 consecutive parts for training and testing, where the training part accounts for $40\%\sim80\%$ of the multi-view sequences, and the novel poses include the remaining part of the videos and other poses like poses from AIST++ dataset~\cite{li2021ai}.

{\bf Metrics.} We report quantitative results using four standard metrics: Peak Signal-to-Noise Ratio(PSNR), Structure Similarity Index Measure(SSIM)~\cite{wang2004image}, Learned Perceptual Image Patch Similarity(LPIPS)~\cite{zhang2018unreasonable} and Frechet Inception Distance(FID)~\cite{heusel2017gans}.

\subsection{Results and Comparisons.}

\hspace{0.422cm}We train the networks for each multi-view video sequence individually. The qualitative results shown in Fig.~\ref{fig:fig1} demonstrate that our approach represents fine-grained dynamic details under various novel poses. For more novel pose synthesis results, please refer to the supplementary materials. Then we compare our method against other approaches, such as TAVA~\cite{li2022tava}, ARAH~\cite{wang2022arah}, AniNeRF~\cite{peng2021animatable}, PoseVocab~\cite{li2023posevocab} and NeuralActor~\cite{liu2021neural}. We do not compare our results with TexDVA~\cite{remelli2022drivable} because TexDVA takes 2 or 3 images to render appearances, which is different from our approach that reconstructs avatar driven by poses.

\begin{figure*}[htb]
  \centering
  \includegraphics[width=1.0\linewidth]{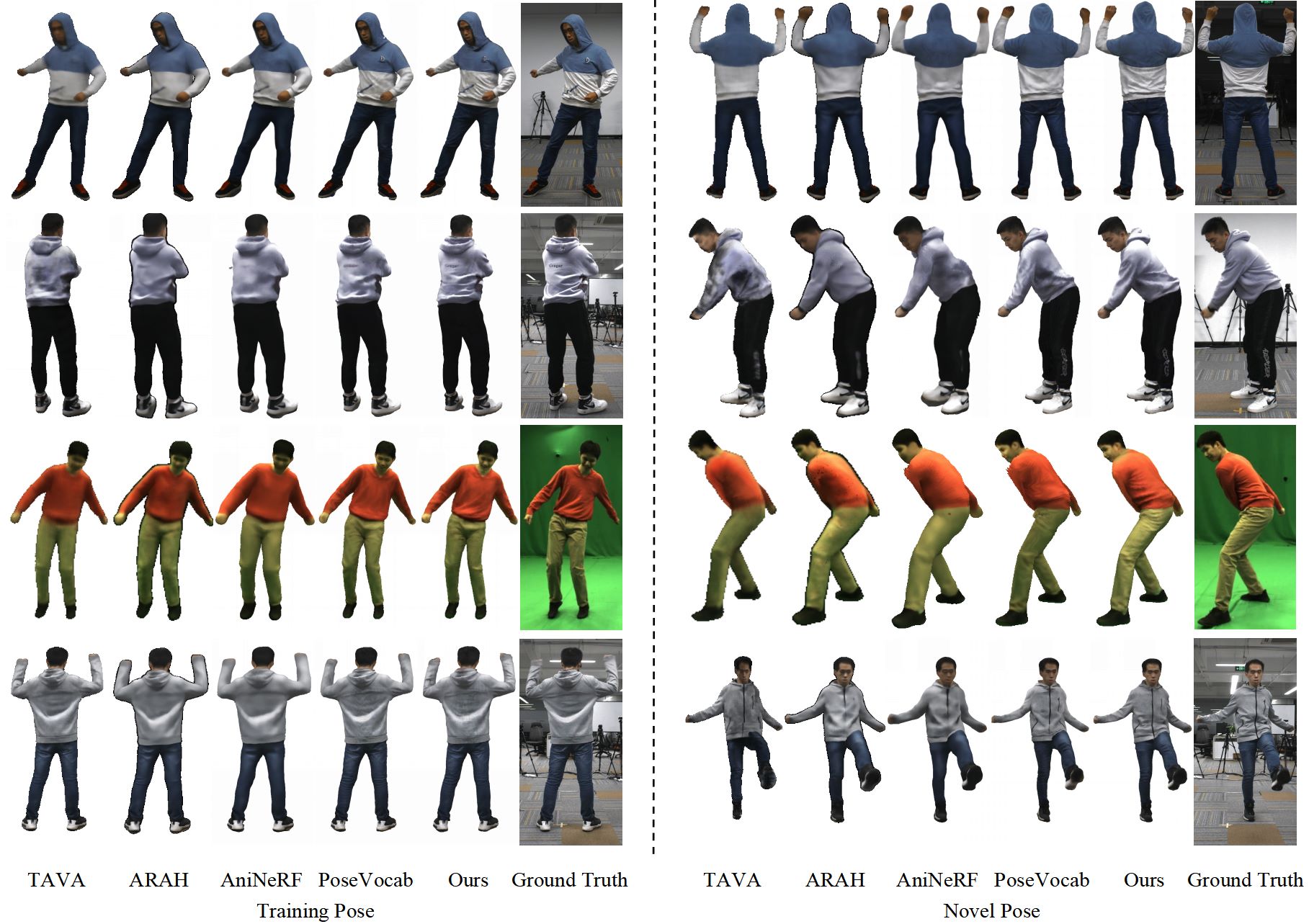}

   \caption{Qualitative comparisons against TAVA, ARAH, AniNeRF and PoseVocab. We evaluate methods on THUman4.0 dataset and DeepCap dataset and show the animation results on both training poses and novel poses respectively.}
   \label{fig:qual-comp}
\end{figure*}

{\bf TAVA, ARAH, AniNeRF and PoseVocab.}
Fig.~\ref{fig:qual-comp} and Tab.~\ref{tab:quan-t4} show the qualitative and quantitative results against TAVA, ARAH, AniNeRF and PoseVocab on DeepCap dataset~\cite{habermann2020deepcap} and THUman4.0 dataset~\cite{zheng2022structured}.
The results synthesized by TAVA, ARAH and AniNeRF are blurry, probably because neither pose-dependent shading in TAVA nor per-frame latent codes in ARAH and AniNeRF provide effective conditions to decode fine-grained details.
Although PoseVocab outperforms the other three approaches, there are still some fine-grained details it cannot represent because of the limited capability of feature lines.
In contrast, our method not only reconstructs more details in terms of the edges and fine-grained garment wrinkles under the training poses, but also renders more realistic dynamics when giving novel poses benefiting from the realistic human appearance conditions served by pixel-aligned pose features.
Overall, our method outperforms the other four approaches both qualitatively and quantitatively with the construction of TexVocab which adequately utilizes the existing image evidence.

\begin{table*}[htb]
\centering
\begin{tabular}{c|cccc|cccc}
\hline
\multirow{2}{*}{Method} & \multicolumn{4}{c|}{Training Poses} & \multicolumn{4}{c}{Novel Poses} \\ \cline{2-9} 
                 & PSNR$\uparrow$ & SSIM$\uparrow$ & LPIP$\downarrow$ & FID$\downarrow$ & PSNR$\uparrow$ & SSIM$\uparrow$ & LPIPS$\downarrow$ & FID$\downarrow$ \\ \hline
               Global Pose  & 33.75 & 0.9824 & 18.65 & 22.59  & 29.59 & 0.9731 & 22.70 & 34.26 \\
                SMPL Joint  & 33.43 & 0.9813 & 19.40 & 21.83  & 29.23 & 0.9732 & 23.54 & 37.58    \\
                Body Part (Ours)  & {\bf34.05} & {\bf0.9878} & {\bf14.51} & {\bf17.30}  & {\bf30.95} & {\bf0.9786} & {\bf19.06} & {\bf28.11} \\
\hline
\end{tabular}
  \caption{Quantitative results of ablation study on encoding strategy. We assign the texture maps to global poses, SMPL joints and body parts respectively, and report the numerical results on 3 sequences of THUman4.0 dataset~\cite{zheng2022structured}. Here LPIPS* = 1000 $\times$ LPIPS. }
  \label{tab:quan-abl}
\end{table*}

\begin{figure}[htb]
  \centering
  \includegraphics[width=1.0\linewidth]{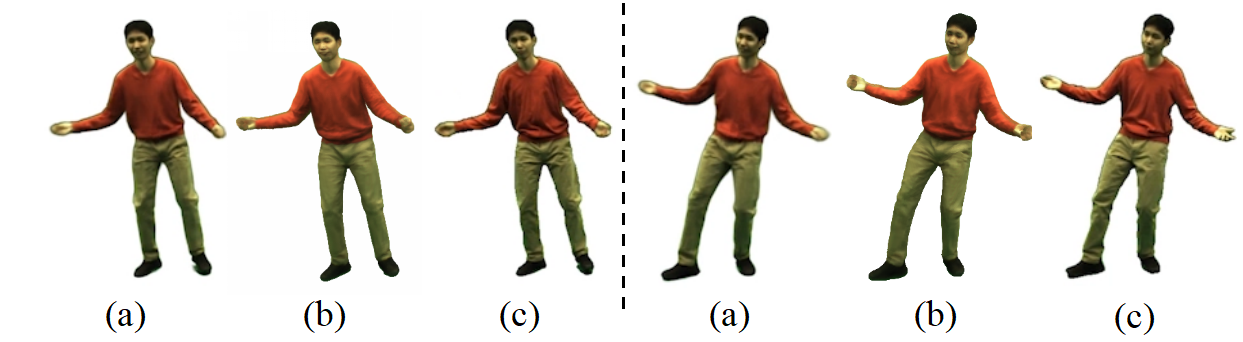}

   \caption{Qualitative comparisons against NeuralActor on novel pose synthesis. Results of NeuralActor are borrowed from~\cite{liu2021neural}. (a) results of NeuralActor, (b) our results, (c) ground truth.}
   \label{fig:fig6}
\end{figure}

{\bf NeuralActor.}
NeuralActor utilizes a 2D vid2vid model~\cite{wang2018video} and tries to predict the SMPL-defined texture maps from normal maps, thus extracting features and using a neural radiance field to decode the 3D human characters.
We compare our method with NeuralActor~\cite{liu2021neural} qualitatively and quantitatively on “S2” sequence of DeepCap dataset in Fig.~\ref{fig:fig6} and Tab.~\ref{tab:quan_351}. We follow the same training/testing splits and metric computation as NeuralActor, and the numerical and visual results are borrowed from~\cite{liu2021neural}.
Although the vid2vid model is powerful enough to predict texture maps solely from SMPL-derived attributes, the low frequency of normal maps still limits the performance. In contrast, directly exploiting texture maps gathered from image evidence can eliminate such problem.

\begin{figure}[htb]
  \centering
  \includegraphics[width=0.98\linewidth]{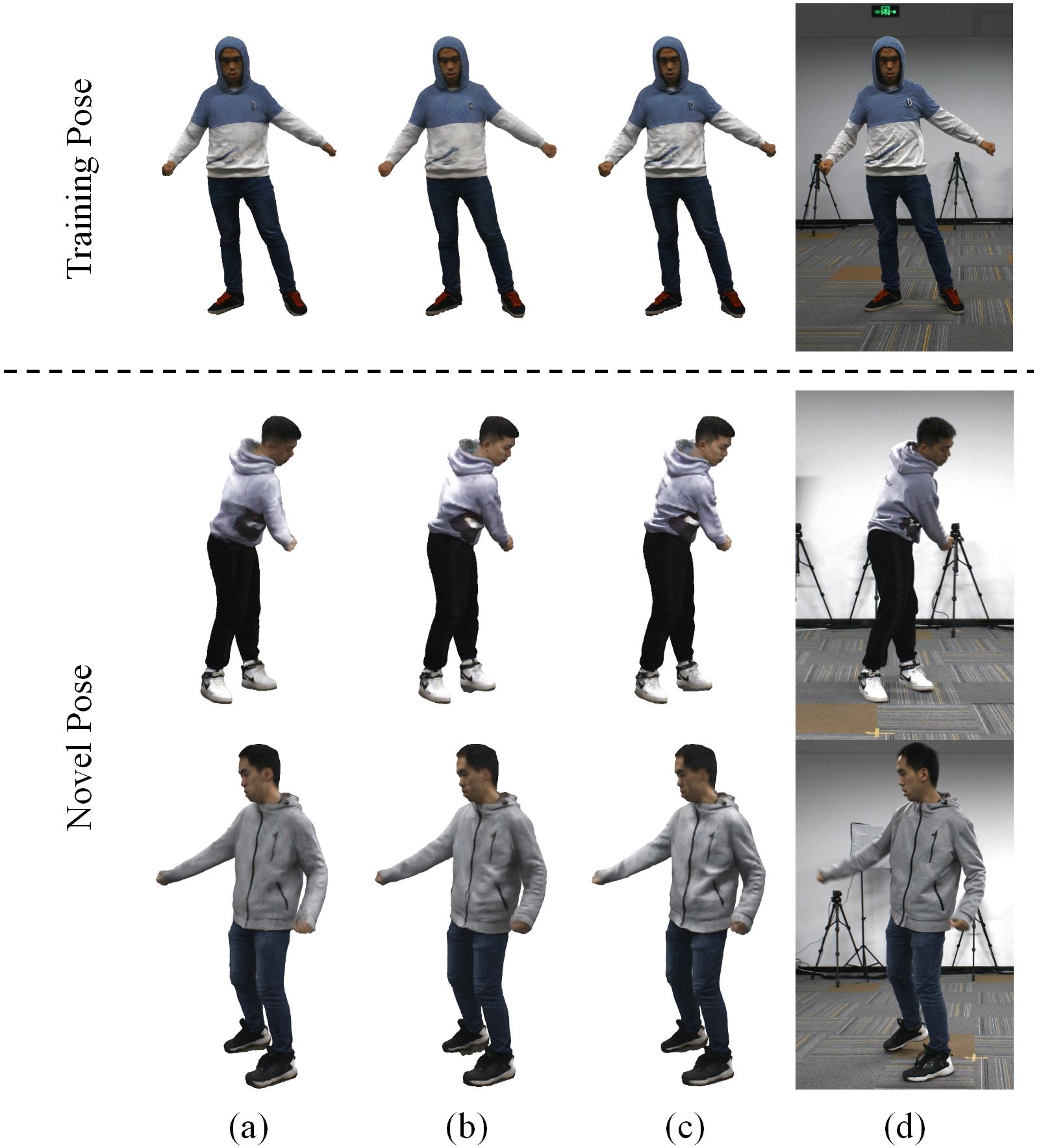}

   \caption{Qualitative results of ablation study on embedding strategy. We show synthesized images of (a)global pose, (b)joint-structured, (c)body-part-wise embedding , and (d)ground truth.}
   \label{fig:fig7}
\end{figure}

\subsection{Ablation Studies}

\hspace{0.422cm}In this subsection, we conduct ablation studies to demonstrate the improvement brought by our contributions.

{\bf Ablation Study on Body-Part-Wise Embedding.} To prove the effectiveness of the body-part-wise embedding, we compare it against another two embedding strategy, \textit{i.e.}, global pose embedding and joint-structured embedding. Specifically, we sample key global poses, key SMPL joints and key body parts and assign texture maps to them, respectively. Fig.~\ref{fig:fig7} and Tab.~\ref{tab:quan-abl} show the qualitative and quantitative results on 3 sequences of THUman4.0 dataset~\cite{zheng2022structured}. Although all the three methods can reconstruct details under training poses benefiting from the pixel-aligned features given by texture maps, global pose embedding cannot disentangle the effects of SMPL joints, and ends up with poor generation. While joint-structured embedding encodes features per joint and does not retain any information on kinematic chains, which makes it too local to generalize to challenging poses, so the wrinkles are often messy. In contrast, the proposed body-part-aware embedding not only encodes fine-grained details, but also displays realistic results by disentangling the effects of different joints and maintaining information along the kinematic chains.

\begin{figure}[tb]
  \centering
  \includegraphics[width=1.0\linewidth]{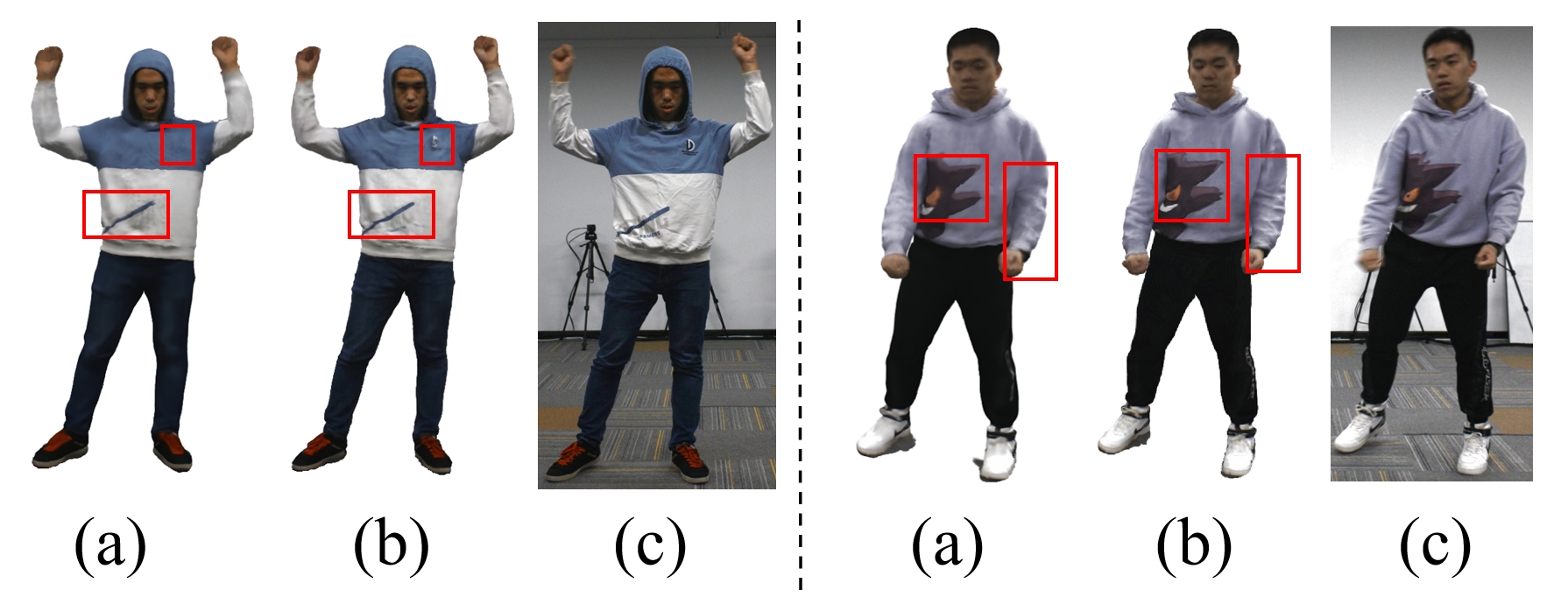}

   \caption{
   Qualitative results on novel pose synthesis of ablation study on texture maps.(a) Texture maps from an image generator, (b) texture maps from multi-view images, (c)ground truth.
   }
   \label{fig:fig8}
\end{figure}

{\bf Ablation Study on Texture Map.} To prove the effectiveness of texture maps obtained from existing views, we 
take SMPL normal map as input, which is similar to NeuralActor~\cite{liu2021neural}. Then we use an image generator to generate texture maps from normal maps, and take ground-truth texture maps as monitoring signals. Fig.~\ref{fig:fig8} shows the qualitative results of novel pose synthesis on THUman4.0 dataset~\cite{zheng2022structured}. Although the 2D convolutional network is powerful enough to generate realistic texture maps from SMPL-derived normal maps, it still suffers from the low representation ability of driving signals. While our method can not only reconstruct fine-grained details but also achieve pose generalization with the guidance of pixel-aligned features.

\section{Discussions}

\hspace{0.422cm}{\bf Conclusion.} We present TexVocab, a texture vocabulary that adequately utilizes explicit image evidence to guide the implicit NeRF to learn the dynamic details from expressive texture conditions. To further represent the dynamic wrinkles of clothed humans, we propose body-part-wise embedding to decompose all the SMPL skeletons into several body parts and assign texture maps to them, which both disentangles the effects of SMPL joints and maintains information of kinematic chains. The proposed approach not only reconstructs fine-grained details in terms of garment wrinkles and edges, but also achieves pose generalization that displays realistic dynamic appearances under novel poses. Experiments on different multi-view video datasets indicate that our approach outperforms other state-of-the-art methods both qualitatively and quantitatively, showing its enormous potential in different kinds of interactive applications.

{\bf Limitation.} Since our avatar representation relies on the inverse skinning by SMPL skeletons, it cannot handle loose clothes like long dresses which do not follow the topological structure of the SMPL model. Also, the gathering of texture maps relies on dense views. For monocular or sparse view datasets, we cannot gather complete texture images to guide avatar construction.

{\bf Potential Social Impact.} Since our method enables automatic creation of animatable human avatars, it can be misused to re-target individuals with actions they do not perform. To prevent the risks, it is critical to evaluate the caution before developing such kind of technology.

{\bf Acknowledgment.} This research was funded through National Key Research and Development Program of China (Project No. 2022YFB36066). The project was also supported by the NSFC project No. 62125107.



\appendix
\setcounter{figure}{0}
\renewcommand{\thefigure}{A\arabic{figure}} 

{
    \small
    \bibliographystyle{ieeenat_fullname}
    \bibliography{main}
}

\end{document}